\pdfoutput=1
\documentclass[11pt,a4paper]{article}
\usepackage[hyperref]{acl2021}
\usepackage{times}
\usepackage{latexsym}

\usepackage{microtype}
\usepackage{multirow}
\usepackage{float}
\usepackage{amsmath}
\usepackage{graphicx}
\usepackage{tabularx}
\usepackage{booktabs}
\usepackage[final]{changes}
\usepackage{pbox}

\newcommand{\contribfootnote}[1]{%
  \begingroup
  \renewcommand{\thefootnote}{}\footnote{#1}%
  \addtocounter{footnote}{-1}%
  \endgroup
}

\newcommand{\fieldheading}[1]{\multicolumn{1}{c}{\textbf{#1}}}
\definechangesauthor[name=KL color=orange]{kl}

\newcommand{\source} {principal}
\newcommand{\ent}{$\langle\mid$ENT$\mid\rangle~$}
\newcommand{\tfidf}{$\langle\mid$TFIDF$\mid\rangle~$}
\newcommand{\citingsent}{explanation}
\newcommand{\citingsents}{explanations}

\aclfinalcopy 
\def\aclpaperid{2108} 

\title{Explaining Relationships Between Scientific Documents}

\author{Kelvin Luu$^{1,*}$
\quad Xinyi Wu$^{1,*}$
\quad Rik Koncel-Kedziorski$^{1}$
\\
\textbf{Kyle Lo}$^{2}$
\quad \textbf{Isabel Cachola}$^{3}$
\quad \textbf{Noah A. Smith}$^{1,2}$ 
\\ 
\textsuperscript{$1$}University of Washington \quad
\textsuperscript{$2$}Allen Institute for AI \quad
\textsuperscript{$3$}Johns Hopkins University \\

\texttt{\{kellu,nasmith\}@cs.washington.edu}, \texttt{\{xywu,kedzior\}@uw.edu},\\
\texttt{kylel@allenai.org},
\texttt{icachola@cs.jhu.edu}}
\date{}

\begin{document}
\maketitle
\begin{abstract}\contribfootnote{$^{*}$Equal contribution.}
We address the task of explaining relationships between two scientific documents using natural language text.
This task requires modeling the complex content of long technical documents, deducing a relationship between these documents, and  expressing that relationship in text.
 Successful solutions can help improve researcher efficiency in search and review. 
In this paper, we operationalize this task by using citing sentences as a proxy. We establish a large dataset for our task. 
We pretrain a large language model to serve as the foundation for autoregressive approaches to the task. 
We explore the impact of taking different views on the two documents, including the use of dense representations extracted with scientific information extraction systems. 
We provide extensive automatic and human evaluations which show the promise of such models, and make clear the challenges for future work. 
\end{abstract}

\section{Introduction}

The output of the world's scientists doubles roughly every nine years \citep{Bornmann_2015}. Consequently, researchers must devote significant energy to quickly understand how a new piece of research fits with a rapidly changing research landscape. 


Several lines of research seek to reduce this burden on scientists. Citation recommendation systems suggest references to relevant published work \citep{McNee2002OnTR,Bhagavatula2018ContentBasedCR}.
Intent classification systems help determine the type and importance of a citation in a work \citep{Valenzuela2015IdentifyingMC,Cohan2019StructuralSF}.
Summarization systems aim to help researchers more quickly understand the basic ideas in a piece of research \citep{Cohan2015ScientificAS,Yasunaga2019ScisummNetAL}. We draw inspiration from these works as well as broader challenges like explaining the connection between concurrent works or relating a new paper to those a reader is already familiar with.


Automatically describing inter-document relationships could decrease the time researchers devote to literature review. For instance, explanations for a new paper can be personalized to a particular reader by relating the new work to ones they have read before. Further, such technology could be incorporated into writing assistance systems to help less experienced or non-native writers better articulate the connection between their work and prior art. Additionally, users of citation recommendation systems can benefit from natural language explanations of recommendation system choices.

\begin{figure}[t!]
    \centering
    \includegraphics[width=0.49\textwidth]{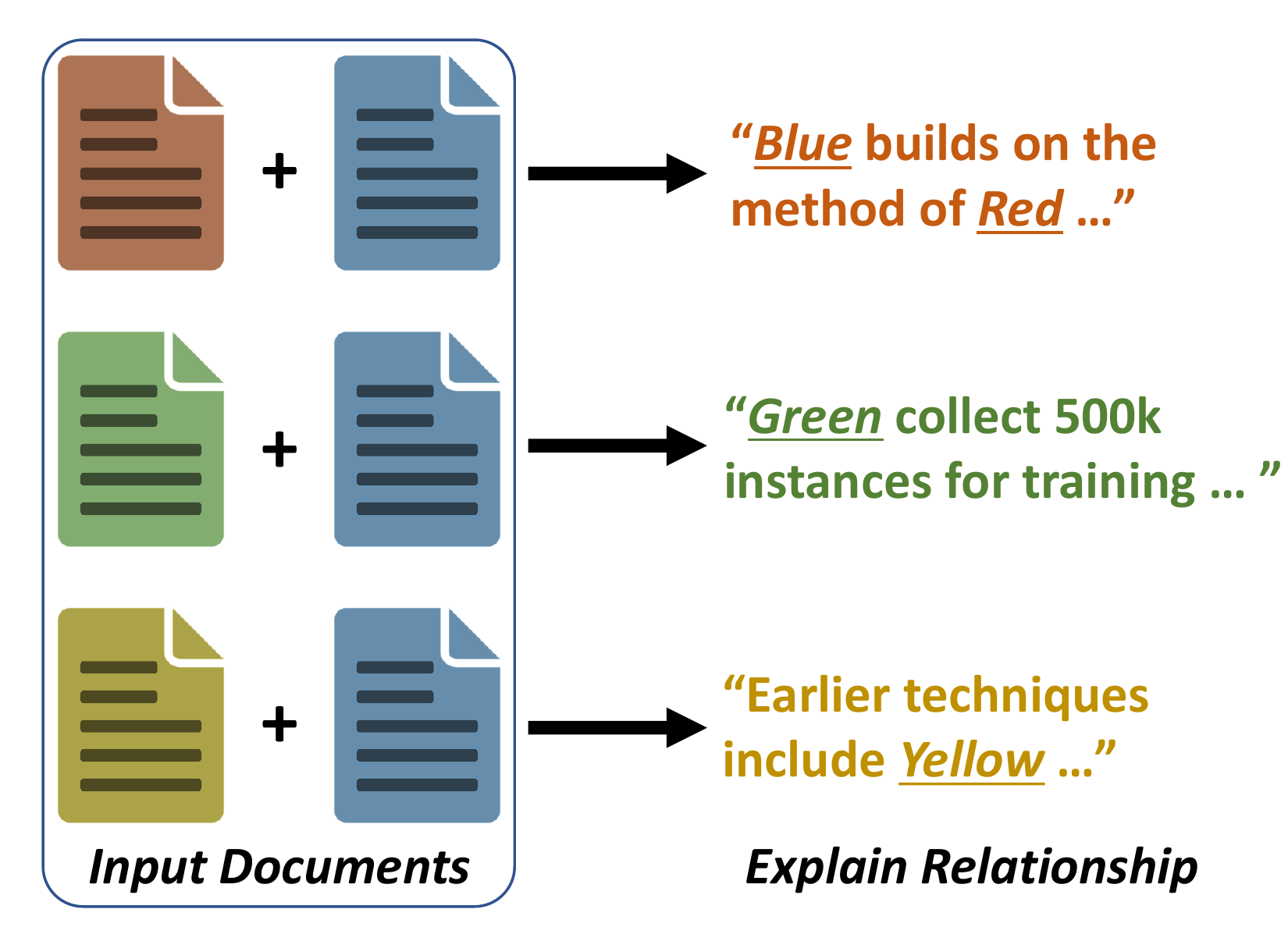}
    \caption{Given two scientific documents, the goal is to write the sentence describing the specific relationship between them. For a given document (in blue above), the output will vary depending the content of the other. (This image is best viewed in color.)}
    \label{fig:teaser}
\end{figure}

In addition to the utility of this task to scientists, it presents several interesting technical challenges. These include effectively representing the important information in a document, generating from a long-tailed technical vocabulary, and expressing the variety of connections between related scientific papers. Figure~\ref{fig:teaser} illustrates how the same document is described differently in relation to different documents. 

In this paper we use citing sentences to operationalize the problem of generating natural language explanations of the relationships between two scientific papers. Authors, when citing other work, oftentimes describe how their work relates to the cited work. To this end, we use in-text citation sentences as a naturally occurring proxy explanations for how two documents relate to each other. 
However, we generate such sentences from general representations of document content rather than the specific in-text locations where these sentences occur, as this task formulation can better facilitate the applications described above.

We approximate the explanation objective by having a GPT2 language model generate sentences containing citations given a pair of documents. This approach relies on providing dense but informative representations of documents to use as conditioning context for the generation model. We explore the use of sentence-based contexts as input including document abstracts, introductions, and sampled sentences from the full document; we find that using introductions and abstracts works well. Finally, we improve our model's performance on automated metrics by using informative entities and terms to both construct dense input and rank the output relationship explanations.


In addition to standard automatic metrics, we perform human evaluations of  technical outputs with a pool of annotators. 
In this work, we describe a series of stages of model development, each with its own experiments that, together, informed the task and our series of solutions. 

Our contributions include:  a novel dataset for the relationship explanation task; a domain-adapted GPT2 we release for left-to-right language modeling of scientific text; the {\sc SciGen} model for describing document relationships; and an extensive expert evaluation and analysis of machine generated technical text.\footnote{\url{https://github.com/Kel-Lu/SciGen}}

\section{Related Work}

The current work builds on recent research in scientific document understanding, including citation recommendation, intent categorization, and scientific document summarization.
Citation recommendation systems suggest related works given a document or a span of text \citep{McNee2002OnTR,Nallapati2008JointLT,Bhagavatula2018ContentBasedCR}.
Recently, researchers have sought to categorize citations using various ontologies of citation intents. \citet{teufel2006automatic} develop an annotation scheme and corresponding classification model for citation functions.
\citet{Valenzuela2015IdentifyingMC} seek to discern ``highly influential'' citations from others.
\citet{jurgens2018citation} use six categories including ``motivation,'' ``uses,'' and ``future work'' among others. \citet{Cohan2019StructuralSF} condense this ontology to just three: ``background,'' ``method,'' and ``result comparison.'' Intent classification can identify relationships between documents; our relationship explanation task extends this in two ways. First, data-driven freeform generation can express a wider array of relationships compared to a manually-defined label set. Further, our task framework could be used to describe relationships between works which do not actually cite each other, such as contemporaneous works. Unlike categorization techniques, we require no task-specific annotated data as we supervise with citing sentences that are readily available in scientific documents. 
In practice, citation classification is used to assist in suggesting relevant works to researchers; our work complements this goal by providing rationales for the recommendation and furthering progress toward explainable AI.

Our work is also connected to a long history of research on summarizing scientific documents \citep{Luhn1958TheAC,Paice1980TheAG}. Work in this area has mostly used used abstracts or peer reviews as targets \cite{Cachola2020TLDRES, Cohan2018ADA, Jaidka2017TheCS}. In particular, \citet{Subramanian2020OnEA} show that using a simple extractive summary as input for abstractive summarization of scholarly texts work well.  Researchers have also used citing sentences as part of the input for summarization, recognizing the explanatory power of these texts
\citep{nakov2004citances,cohan2017contextualizing,Yasunaga2019ScisummNetAL}. Ours is the first work to focus on learning to express the specific relationship between two documents from such sentences. 

The closest work to our own is \citet{xing-etal-2020-automatic}, who pilot a task of in-line citation generation. Their goal is a model which can insert a citing sentence into a particular context within a document. 
Our work, on the other hand, aims to learn from citing sentences how to describe general relationships between documents independent of particular in-document contexts. 
While the \citet{xing-etal-2020-automatic} method may facilitate writing assistance, our task has applications in search and summarization. 
Because our task does not rely on a specific location in a document where the citation will go, solutions can be used at scale to provide users with general explanations of document relationships.

Our models rely heavily on recent advances in transfer learning in NLP. 
Large pretrained models such as BERT \citep{devlin2018bert} and GPT2 \citep{radford2019language} have made strong advances on a number of tasks \citep{wang2019superglue}.
It has also been shown that pretraining these models on domain-specific data further improves results on domain-specific tasks \citep{Beltagy2019SciBERT,Lee2019BioBERTAP}. 
In this work, we apply that methodology by adding a pretraining phase on in-domain data before finetuning a GPT2 model toward the explanation generation task.
A key challenge when using pretrained language models for document-level tasks is how to select document content to fit within the limited context window of the model, which is a major focus of our work.

\section{Task Overview}

We aim to generate an \citingsent{}: a natural language sentence which expresses how one document relates to another. Explicit examples of such sentences are nontrivial to find in corpora, especially when annotation for a highly technical task is expensive. To this end, we use in-text citations in a scientific document to prior work as proxies for relationship explanations. We use these citing sentences as partial supervision for our task, and refer to them as ``explanations.''\footnote{Future work might seek to filter or systematically alter in-text citations to be more explanation-like, without otherwise changing our approach.} 

We distinguish one document as the {\it principal} document, from which we will draw \citingsents{} that reference the {\it cited} document.
Let $t$ denote an \citingsent{} drawn from principal document $S$, and $S'$ denote $S$ without $t$. Then let
\begin{equation}
    P(t\mid S',C)
    \label{eq:prob}
\end{equation}
be the probability of  $t$ given $S'$ and the cited document $C$. 
A good generation technique should maximize this probability across a large number of $\langle t,S,C \rangle$ triples, so that at inference time the model is able to generate a sentence $t^\ast$ which accurately describes the relationship between new documents $\hat{S}$ and $\hat{C}$.

Optimizing Equation~\ref{eq:prob} is made easier by modern representation learning.
Pretrained neural language models like GPT2 have  shown strong performance when generating sentences conditioned on a context.
However, existing implementations of GPT2 limit the context window to 512 or 1024 tokens, far smaller than scientific documents.
In this work, we explore ways to represent the documents' content for use with language models. 


\paragraph{Data} \label{sec:data}We use English-language computer science articles and annotation from S2ORC dataset \citep{lo-wang-2020-s2orc}. 
S2ORC is a large citation graph which includes full texts of 8.1 million scientific documents. 
We use 154K connected computer science articles, from which we extract 622K \citingsents{} with a single reference that link back to other documents in our corpus. We omit any sentences that cite more than one reference.
We hold 5000 sentences for each of the validation and test sets.
Detailed statistics can be found in Table~\ref{tab:dataset}. Information on dataset construction can be found in Appendix~\ref{appendix:dataset}. 


\begin{table}[t]
    \centering
    \begin{tabular}{l|r|r}
         & total & average/doc.  \\
        documents & 154K & -- \\
        tokens & 813M & 5.3K  \\
        unique tokens & 7.1M & 1.3K  \\
        \citingsents{} & 622K & 4.0  
    \end{tabular}
    \caption{Dataset statistics, total and per document.}
    \label{tab:dataset}
\end{table}

\paragraph{Evaluation} The most appropriate evaluation metric for this and many text generation tasks is human judgment by potential users of the system. 
Evaluating explanations of the relationships between scientific documents requires human judges with scientific expertise whose time and effort can be costly. 
While collecting human judgments in technical domains is relatively rare, we believe it to be an important step in evaluating our systems for this task.  Thus, we conduct thorough human evaluations and analyses with expert judges.
We make use of both larger scale expert evaluations yielding hundreds of judgements as well as smaller scale, deeper evaluations where we can effect a higher degree of quality control over fewer datapoints. 
Further, we make use of intermediate human evaluations in the development of our models, and supplement these evaluations with automatic metrics ---  BLEU \citep{papineni2002bleu} and ROUGE \citep{Lin2004ROUGEAP} that are established in other generation tasks.

\section{Models}

We develop several 
models for explaining document relationships. 
Following current work in neural text generation, we finetune the predictions of a large pretrained language model to our task (Section~\ref{sec:ntg}). 
In order to bring the language model into the scientific text domain, we do additional language model pretraining over full scientific texts.
We also investigate approximate nearest neighbor methods to retrieve plausible human-authored \citingsents{} from the training data as a baseline (Section~\ref{sec:ir}).

\subsection{Neural Text Generation}



\label{sec:ntg} 
Recent work has shown that finetuning large pretrained language models to text generation tasks yields strong results \citep{zellers2019neuralfakenews}. To this end, we construct {\sc SciGen}, a model based on GPT2 \citep{radford2019language}, a transformer model trained on 40GB of internet text with a left-to-right language modeling objective \citep{vaswani2017attention}. We do so by finetuning the predictions of the language model to generate explanations using different expressions of the principal and cited document as context. 


To finetune GPT2 architectures for text generation, it is typical to concatenate the conditioning context $X = x_1 \ldots x_n$ and target sentence $Y = y_1 \ldots y_m$ with a special separator token $\xi^y$. To adapt this technique to our task, we construct the conditioning context $X$ from the principal and cited documents and use the \citingsent{} as $Y$. 
We take $j$ tokens from principal document $s_1,\ldots,s_j$ along with $k$ tokens from the cited document $c_1,\ldots,c_k$ (which tokens to draw from the two documents is an independent variable that we explore experimentally).
We then condition the generation of \citingsent{} $Y$ on $X = s_1,\ldots,s_j,\xi^x,c_1,\ldots,c_k$, where $\xi^x$ is a token used to indicate the end of the principal document. {\sc SciGen} is trained to predict the \citingsent{} one token at a time as described above. More details on training can be found in Appendix~\ref{appendix:training}.

At inference time, the model is provided with an unseen principal/cited document pair. 
An explanation of their relationship is generated one token at a time using nucleus sampling \cite{Holtzman2020TheCC}. 
At timestep $t$, output token $\hat{y}_t$ is sampled from the top 90\% of the distribution $P(\hat{y}_t \mid X,\xi^y,\hat{y}_1,\ldots,\hat{y}_{t-1})$ (renormalized).
The selected $\hat{y}_t$ is used to condition the prediction of subsequent tokens.

\paragraph{Context}  The primary question we investigate with the {\sc SciGen} model is what kind of input is best for describing the relationship between the principal and cited documents accurately and informatively. Since models based on GPT2 have a small context window relative to the length of scientific documents, we investigate the use of abstracts, introductions, or non-citing sentences sampled from throughout the document as conditioning context.
The effectiveness and description of these approaches is described in Section~\ref{sec:continuous}.
Based on our findings with sentence-based contexts and information retrieval systems, we then explore the possibility of representing the cited document text as a list of important concepts rather than fluent text, in Section~\ref{sec:terms}.

\paragraph{Language Model Pretraining}
\label{sec:pret}

\begin{figure}[]
    \centering
    \includegraphics[width=0.45\textwidth]{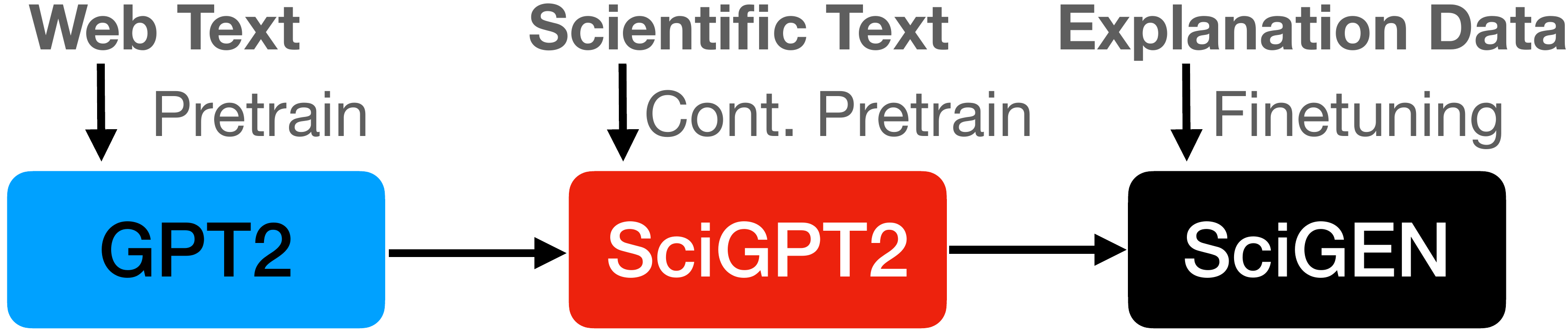}
    \caption{Overview of the construction of {\sc SciGen}. We take the pretrained GPT2 and continue pretraining on scientific texts. We then finetune using data in Table~\ref{tab:dataset}.}
    \label{fig:scigenmodels} 
\end{figure}

Prior work has shown that pretraining on in-domain data improves the performance of large language models on domain-specific tasks \citep{Beltagy2019SciBERT,Gururangan2020DontSP}.
Inspired by this, we continue pretraining the GPT2 model in the science domain to produce {\sc SciGPT2}, which we use as the underlying language model for {\sc SciGen} described above.
{\sc SciGPT2} starts from the standard pretrained GPT2-base model and is trained for an additional 75k gradient updates at batch size of 64 (effectively a single epoch over 4.8 million abstracts and body paragraphs) with a language modeling objective. Figure~\ref{fig:scigenmodels} illustrates the process.

We observed significant improvements in the quality of {\sc SciGen} outputs after replacing the underlying GPT2 language model with the domain-specific {\sc SciGPT2} model. We saw a perplexity improvement in a held-out set and, in informal inspections, qualitative improvements as well. 

When using pretrained language models, text from task-specific test data cannot be guaranteed to be absent from the large task-independent corpora upon which these models are trained, which may improve model performance compared to models without this exposure. 
For the experiments described in this work, we train a version of {\sc SciGPT2} only on documents appearing in the training data, so that the principal documents and target sentences in the test data are unseen by the language model. 
We provide this and a full-corpus version of {\sc SciGPT2} as resources for future research.\footnote{\url{https://github.com/Kel-Lu/SciGen}}


\subsection{Retrieval with Approximate Nearest Neighbors} 
\label{sec:ir}
While neural text generation techniques have advanced significantly in recent years, their outputs are still inferior to human authored texts.
For some tasks, it is better to retrieve a relevant human-authored text than to generate novel text automatically \citep{Fan2018HierarchicalNS}. 
Is this also the case when generating explanations?

To answer this question, we use an information retrieval (IR) baseline. We adapt an approximate nearest neighbor search algorithm to find similar pairs of documents. 
The basic search procedure is as follows:
Given a test instance input $(S,C)$ for \source~$S$ and cited document $C$, we find the set $\mathbf{N}_C$, the nearest neighbors to $C$ in the training data.
For each document $N_C$ from  $\mathbf{N}_C$, let $\mathbf{N}_S$ be the set of documents that cite $N_C$.
This means that each $N_S \in {\mathbf N}_S$ contains at least one citing sentence $t'$ which cites $N_C$. 
We use the $t'$ associated with the $(N_S,N_C)$ pair from the training which is closest to $(S,C)$ as the explanation of their relationship. 

We measure the closeness of two pairs of documents using the cosine distances between vector representations of their abstracts.
The abstract of each document is encoded as a single dense vector by averaging the contextualized embeddings provided by the SciBERT model of \citet{Beltagy2019SciBERT} and normalizing.
The distance between $(S,C)$ and neighbors $(N_S,N_C)$ is computed as:
\begin{equation}
    \alpha\cos(S,N_S)+\beta\cos(C,N_C)
\end{equation}
where $\alpha$ and $\beta$ control the relative contribution of the two document similarities. 
We explore setting both $\alpha$ and $\beta$ to 1, or tuning them to optimize BLEU on the validation data using MERT \citep{och2003minimum}.
\section{Representing Documents with Sentence Selection}
\label{sec:continuous}

Methods for the related task of citation recommendation have made use of abstracts, which perhaps act as sufficient summaries of document content.
Building on this, we represent the principal and cited documents with the first 450 tokens of either their
{abstracts},  {introductions}, or  sentences randomly sampled from throughout the full document.\footnote{\label{footn:fullsample}We exclude any sentence with a citation from being sampled in all conditions. This context type is also only used for the cited document and not the \source~document.} In this section, we answer two questions: 1) do neural generation models with sentence-based context outperform the IR baseline and 2) does the type of sentence-based context (abstract, introduction, sampled) matter? 
We answer these questions by performing both automatic and human evaluations.

\subsection{Automatic Evaluation}

We compare the {\sc SciGen} and IR systems using BLEU \citep{papineni2002bleu} and ROUGE  (specifically L; \citealp{Lin2004ROUGEAP}).
The ``Sentence-based''  rows of Table~\ref{tab:autometrics} show the test set performance of the IR system and the best {\sc SciGen} models when provided with the different sentence-based input context combinations.\footnote{The performance of our best {\sc SciGen} models can be found in Table~\ref{tab:autometrics} and the automatic test set evaluations of all systems can be found in  
Appendix~\ref{appendix:autometrics}.}
We assesss statistical significance as well by bootstrapping with 1000 samples in each of 100 iterations. We find that context \emph{does} make a difference for {\sc SciGen}, and that a slight but statistically significant performance improvement comes from using the introduction of the \source~document rather than the abstract.\footnote{$p<0.01$ after Bonferroni correction.} We do not, however, find enough evidence to reject the null hypothesis that any particular representation of the cited document's content (abstract, intro, or random sample) is sufficient. 

We find that using the introduction of the principal document paired with the abstract of the cited document performs best, and so we select these for human evaluation. \comment[id=kl]{removed "abstract or intro". Also added stat tests. Does it fit here?}
The IR systems perform well, obtaining slightly better scores in some settings. We choose the MERT-optimized version for human evaluation.


\begin{table}[t]
    \centering
    \begin{tabular}{lccc|c}
        \toprule
         & {\bf Specific} & {\bf Correct} & {\bf S\&C} & {\it agr}\\\hline
        {\sc SciGen} & {72.3} & {64.0} & {55.0} & {\it 70.5}\\
        {IR}  & {74.8} &  {46.3} & {40.0} & {\it 77.5}\\
        Gold & 81.4 &  72.1 & {68.0} & {\it 83.8 }\\\hline
        {\it agreement }& {\it 69.8} & {\it 71.4 } & {\it 63.1}\\
        \bottomrule 
    \end{tabular}
    \caption{ Human evaluation of {\sc SciGen} (intro $\times$ abs) and IR (abs $\times$ abs) systems compared with gold \citingsents{} in percent. S\&C represents those that were both specific and correct. All differences significant at $p<0.01$ except {\sc SciGen} vs.~IR specific.}
    \label{tab:human1}
\end{table}

\subsection{Human Evaluation}

\label{subsec:human1}
We conduct a human evaluation to determine, given a particular pair of \source~and cited abstracts, how {\it correct} and {\it specific} the generated explanation of their relationship is.
By ``correct'' we mean: does the explanation correctly express the factual relationship between the \source~and cited documents?
Because generic explanations such as ``This work extends the ideas of Chomsky and Halle (1968)'', while possibly factual, do not express a detailed understanding of the documents' relationship, we ask judges whether the explanation describes a specific relationship between the two works. 
An explanation can be specific even it is incorrect.



We compare the {\it \source~intro $\times$ cited abs} {\sc SciGen} setting against the tuned IR system. 
For calibration, we also elicit judgments for the gold \citingsents{} extracted from \source~documents along with the correct \source~and cited abstracts. 
In all three cases, we ensure that the \source~document appeared in the ACL anthology to ensure annotator expertise. 
In total we solicit 37 NLP researchers and collect over 800 judgments, with over 100 for each system/quality dimension combination. 

Further details of our evaluation can be found in Appendix~\ref{appendix:humaneval}. We perform error analysis on these judgments as well as an additional study to validate human judgments; these are detailed in   Appendix~\ref{appendix:validity} and Appendix~\ref{appendix:error}.
Table~\ref{tab:human1} shows the percentage of ``yes'' judgments versus the total of ``yes'' and ``no'' judgements for each system/quality combination, along with pairwise agreement rates.\footnote{That gold texts do not achieve perfect scores demonstrates a limitation of our evaluation setup, due in part to the fact that judgments are based on document abstracts rather than their full texts. We take steps to resolve this limitation in our subsequent analysis in Section~\ref{subsec:human2}.} Gold texts received the highest scores for all dimensions of text quality from the evaluators as well as the highest agreement rate. 
We can also see that IR systems tend to produce incorrect explanations more often than not.

The {\sc SciGen} system performs quite well in this analysis, with a majority of outputs deemed correct. 
We observe a larger difference in specificity between {\sc SciGen} and gold texts, indicating that {\sc SciGen}, like many neural text generation systems, often generates vague and generic sentences. These generations tended to be vacuous such as ``\textsc{(Cited)} This work is an extension of the paper.'' Specificity is key for future downstream applications such as automated literature review and will need to be improved for those tasks.

\section{Using IE-Extracted Term Lists}
\label{sec:terms}



Compared to the gold \citingsents{}, we found that our generated explanations miss important phrases such as unique model or dataset names and other lower-frequency terms; generally, they lacked specificity. 
The missing phrases typically appear in the cited document after the abstract and introduction.\footnote{A quantitative analysis of this phenomenon is available in Appendix~\ref{appendix:token}.}
Na\"ively sampling from the full text does not capture them due to sparsity. 

\begin{figure*}[]
    \centering
    \includegraphics[width=0.9\textwidth]{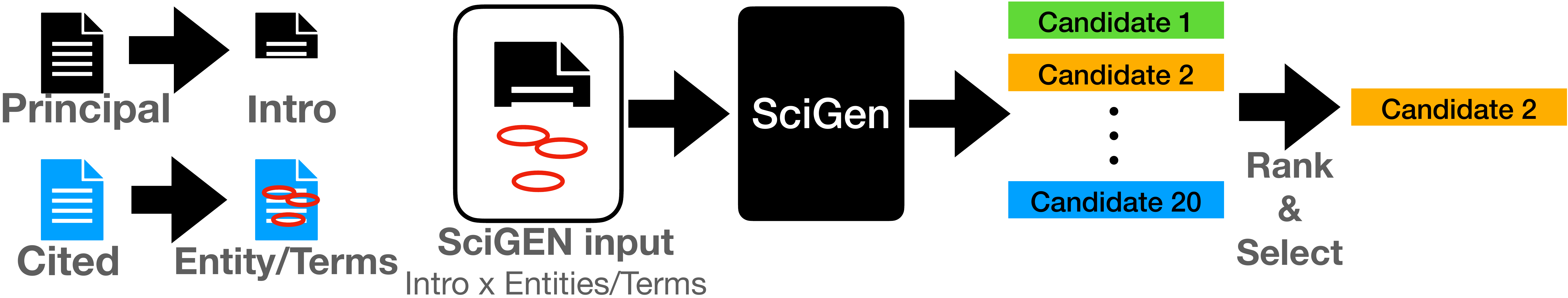}
    \caption{Overview of {\sc SciGEN} using terms/entities. We generate a list of candidates and rank them according to mean reciprocal rank to the input entities. } 
    \label{fig:inputs} 
\end{figure*}

To address this issue, we explore more sophisticated information extraction (IE) techniques for constructing the conditioning context for {\sc SciGen}.
Recent work has shown that pretrained language models can adapt to disfluent inputs such as linearized trees and graphs \citep{ribeiro2020investigating}. 
Inspired by this, we investigate whether we can use lists of salient words and phrases to effect a dense representation of the cited document in the conditioning context. 
Specifically, we construct a list of document-specific terms using tf-idf to score unigrams and entities extracted with a state-of-the-art scientific NER system. The paradigm is illustrated in Figure~\ref{fig:inputs}.

\paragraph{Tf-idf} 
Tf-idf is a measure of the frequency of a term in a document, normalized by the document frequency of that term. In our use, we calculate the tf-idf score for each unigram in the cited document. 
We keep the 100 highest scoring terms $w_i$ sorted in descending order of scores. 
The terms of this list are concatenated with a special token $\xi^{\mathit{tf}}$ to signal that this part of the input is structured as a list rather than conventional text.
The resulting context $X^{\mathit{tf}} = w_1, \xi^{\mathit{tf}}, w_2, \xi^{\mathit{tf}}, ... ,\xi^{\mathit{tf}},w_{100}$ is used to represent the cited document to the {\sc SciGen} model.

\paragraph{Entities}
We extract entities from abstracts with the DyGIE++ information extraction framework \citep{Wadden2019EntityRA} using the model trained on SciERC \citep{luan2018multi}, a dataset of scientific document abstracts with entity and relation annotations.\footnote{We found relation annotations to be noisy on inspection.}
The extracted entities $e_i$ from the cited document are sorted by their tf-idf scores compared to all entities in the corpus. 
As above, a special token $\xi^{e}$ is used to concatenate entities and help the language model distinguish this list from conventional text.
If there is additional room in the context window we append the unigrams with the highest tf-idf to the end of the listed entities until the window is full. 
In that case, the cited document context $X^e$ is $e_1, \xi^{e}, e_2, ... , \xi^{e}, e_n,\xi^{\mathit{tf}},w_1\xi^{\mathit{tf}},...,w_m$, where $n$ is the number of entities and $m$ is $100-n$.

\begin{table*}
\centering
\small
\begin{tabular}{lllccc}
\hline
&
\fieldheading{Method} & \fieldheading{Context}& \fieldheading{BLEU} & \fieldheading{ACL-BLEU} & \fieldheading{Rouge-L}\\
\hline
\multirow{6}{*}{Sentence-based} & \multirow{4}{*}{{\sc SciGen}} & \source~abs $\times$ cited abs &  9.82 & 10.40& 8.4\\
& & \source~intro $\times$ cited abs &  9.92 & 11.22 & 8.7\\
& & \source~intro $\times$ cited intro &  9.80 & 10.54 & 8.8\\
& & \source~intro $\times$ cited sampled &  9.81 & 10.31 & 8.7\\
\cline{2-6}
& \multirow{2}{*}{IR} & source abs $\times$ cited abs & 9.93 & 10.50 & 9.7\\
& & $\quad$+MERT & 10.23 &10.29 & 9.8\\
\hline
\multirow{4}{*}{IE-based} & \multirow{2}{*}{{\sc SciGen}} 
 & \source~intro $\times$ cited tfidf & 13.17 & 16.75 & 12.0\\
& & \source~intro $\times$ cited entities & 13.41 & 13.42 & 11.8\\
\cline{2-6}
& \multirow{2}{*}{\shortstack{+Ranking}} 
 & \source~intro $\times$ cited tfidf & 13.50 & 15.10 & 12.3\\
& & \source~intro $\times$ cited entities & 13.16 & 14.47 & 11.8\\
\hline
\end{tabular}
\caption{Automatic test set evaluation of generated texts for a subset of our systems. ACL-BLEU denotes the BLEU scores of the subset of examples we use for human evaluation (see Section \ref{subsec:human1}). The full results can be found in Appendix \ref{appendix:autometrics}.}
\label{tab:autometrics}
\end{table*}

\subsection{Entity-Based Ranking}
\label{sec:fact}
\citet{maynez2020faithfulness} point out that summarization systems frequently struggle with factuality and generate hallucinations unfaithful to input documents.
We observe this problem with some generated explanations as well: popular, topical terms like `CNN' would appear in explanations of papers using LSTM models, for example.  
To combat hallucinations and promote factual accuracy we include a ranking mechanism that rewards generated explanations with higher coverage of important entities from the conditioning context.\footnote{An oracle ranking is shown in Appendix~\ref{appendix:oracle}.}

The process we use is as follows: first, we generate a large space of candidate explanations for a given input document pair from {\sc SciGen} via nucleus sampling. 
We then extract the entities from each candidate using the DyGIE++ IE system. 
Where possible, we match entities from the candidates with the entities extracted from the cited document. 
To account for textual variation between the explanations and the input documents, we use a similarity threshold to make soft alignments.\footnote{We use difflib and a 0.7 similarity threshold for matching.}
We then select the candidate that has the highest mean reciprocal rank of matched entities against the input as the explanation for this document pair.


\subsection{Manual Analysis}
\label{subsec:human2}
We conducted a manual correctness analysis of the generated explanations from a sentence-based (intro $\times$ abs) and IE-based (intro $\times$ tfidf generate and rank) model. 
Two of the authors judged 50 datapoints from each system using a similar setup to that described in Section~\ref{subsec:human1}, but with the single objective of judging correctness on a 3-way scale: Correct; Too Vague (but not incorrect); and Incorrect.
Additionally, the authors made use of the full text of the input documents to make decisions for cases where not enough information is available in the abstract.  
This resulted in a more accurate though much more time-consuming evaluation process compared to the previous evaluation. 
After judging all datapoints independently, the two authors discussed disagreements until a consensus was reached. 

The results of this analysis are shown in Table~\ref{tab:human2}.
We see a slight increase in correctness with the IE-based model compared to the sentence-based model, though the difference is small.

\begin{table}[t]
    \centering
    \begin{tabular}{l|ccc}
    & Correct & Vague & Incorrect \\ \hline
    Sentence-based & 11 & 7 & 32 \\
    IE-based & 13 & 6 & 31 
    \end{tabular}
    \caption{Results of Manual Analysis }
    \label{tab:human2}
\end{table}

\subsection{Automatic Evaluation}
The ``IE-based'' rows of Table~\ref{tab:autometrics} show the results of automatic metrics for the systems described in this Section. 
We find that these metrics improve significantly in the settings where the principal document is represented by its introduction and the cited document is represented either as a list of terms or entities, with a slight advantage for entities.
The models conditioned on intro $\times$ tfidf context outperform all other sentence-based, retrieval, and IE-based models. 

\section{Discussion}
\label{sec:analysis}

\begin{table}[ht]
    \centering
    \small
    \begin{tabularx}{\columnwidth}{llX}
    \toprule 
&Principal: & \emph{A Syllable-based Name Transliteration System} \\ 
1&Cited: & \emph{A Joint Source-Channel Model for Machine Transliteration}\\
&{\sc SciGen}: &Following \emph{Cited} , Chinese characters are considered as Pinyin sequence.\\
\midrule
&{Principal}: & \emph{Recovering discourse relations: Varying influence of discourse adverbials}  \\
2&Cited: & \emph{The Benefits of a Model of Annotation}\\
&{\sc SciGen}: & The two text collections provided by \emph{Cited} were used for training, and the other two text collections were used for evaluation. \\
\midrule
&Principal: & \emph{Coreference Resolution for Swedish and German using Distant Supervision}  \\
3&Cited: & \emph{Collective Cross-Document Relation Extraction Without Labelled Data}\\
&{\sc SciGen}: &  It is one of the most widely used distant supervision techniques and is inspired by techniques proposed by \emph{Cited}. \\
\midrule
&Principal: & \emph{Neural Text Generation in Stories Using Entity Representations as Context}  \\
4&Cited: & \emph{Delete, Retrieve, Generate: A Simple Approach to Sentiment and Style Transfer}\\
&{\sc SciGen}: & The authors of \emph{Cited} proposed a model that combines neural generation with user interaction to create an object-centric reading experience. \\
\bottomrule
\end{tabularx}
\caption{Example explanations. The given texts are the document titles and the {\sc SciGen} outputs. In the last example, the two documents \textbf{do not} cite each other.}
\label{tab:selectex}
\end{table}

Example system outputs for selected test datapoints are shown in Table~\ref{tab:selectex}. 
The first example illustrates a case where the model identifies a correct relationship between the two documents. 
In this instance, they both use the pinyin representation for Chinese characters in their transliteration models. 

Output 2 demonstrates a failure of the explanation generation system.
The principal document deals with the topic of discourse relations, the automatic identification of which is a long-standing machine learning task. 
However, this particular document is an analysis paper, and does not involve any training. 

Output 3 is an example of a ``Too Vague (but not incorrect)'' case from the analysis in Section~\ref{subsec:human2}.
Here again the explanation generated by {\sc SciGen} is topical, dealing with the concept of ``distant supervision'' that is key to both input documents. 
However, this sentence fails to capture the specific use that the principal makes of the research described in cited document. 

The final example, output 4, showcases potential for our system to explain concurrent work. The generated text summarizes the \emph{cited} and implies that \emph{principal} will build on that work. However, selected papers are both concurrent generation papers published in the same venue and do not cite each other. This appears to be a weakness in using citation sentences as proxies for relationship explanations. Citations of contemporaneous work occur less frequently, so these types of sentences appear less often in training.
Similarly, relationship explanations between papers with more distant connections (e.g., ``multi-hop'' in the citation graph) are  missing in our training data. 

In addition to missing some relationships, not all citation sentences are useful as explanations. As pointed out by other work, citation sentences can often be simple summaries of the cited work \cite{qazvinian-radev-2008-scientific, cohan2017contextualizing}. Alternatively, they can be too specific to be useful, as seen in Output 1, where a higher-level summary might be more useful. Future work could focus on curating better training sets for our task. \comment[id=kl]{Added paragraph here to discuss diff bt citations and explanations a bit.}

It is notable that the {\sc SciGen} model usually outputs syntactically correct and topical \citingsents, even given the difficulty of the vocabulary in this domain. 
This is consistent with many recent findings using domain-specific language models. 




The fluency and appropriateness of {\sc SciGen}'s generations shows the promise of generating explanations which accurately capture the relationship between two documents. 
Based on the results obtained here, we expect pretrained scientific language models to persist as a foundation.  
Future work should focus on two complementary goals: ensuring the factual accuracy of the generated text and improved modeling of the cited document. 
Factual accuracy is difficult to enforce in language model-based text generation systems, especially where inference includes sampling procedures. The use of information extraction for contexts showed promise in Section~\ref{sec:terms}; other methods of incorporating information like grounding to knowledge bases could help prune false or irrelevant statements.


Combining knowledge graphs with language models and generation is an active research area that has shown promise in other domains \cite{Bosselut2019COMETCT, koncel2019text, peters-etal-2019-knowledge}. Applying this line of work to scientific text by modeling input documents as knowledge graphs of their content may help algorithms better understand the cited document, provide distant supervision for concurrent work, and result in better outputs. 
\section{Conclusion}
We have described a task of explaining the relationship between two scientific texts and its connections to facilitating researcher productivity. 
We employ a large, publicly available dataset of scientific documents to train a domain-adapted left-to-right language model for use in text generation applications and beyond. 
We explore a collection of techniques for representing document content including using abstracts, introductions, sampled sentences, and lists of informative terms and entities. 
We conduct thorough human and automatic evaluations to determine the relative strengths of each representation for expressing document relationships in natural language text. 
\paragraph{Acknowledgements}
This research was supported in part by the Office of Naval Research under the MURI grant N00014-18-1-2670.
We thank the members of Noah Smith's (ARK), Hanna Hajishirzi's (H2Lab), and Luke Zettlemoyer's research groups for their participation in our study. We thank members of Noah's ARK for their helpful comments and the anonymous reviewers for their feedback. 

\paragraph{Ethical considerations}
The authors received an IRB approval for this human annotation project. The project was classified as ``no-risk.'' All participants in the study were volunteers and gave explicit, informed consent for the study.

\bibliographystyle{acl_natbib}
\bibliography{anthology,acl2021}

\clearpage
\appendix
\label{appendix}
\section{Training Details}
\label{appendix:training}

We perform task-adaptive (continued) pretraining and then finetuning on the task to construct {\sc SciGPT2} and {\sc SciGen} respectively. {\sc SciGPT2} starts from the standard pretrained GPT2-base model and is trained for an additional 75k steps at batch size of 64 (effectively a single epoch over 4.8 million abstracts and body paragraphs) with a language modeling objective. We then finetune {\sc SciGPT2} to build {\sc SciGen} for various contexts. For all variants, we finetune the underlying language model for an additional 10 epochs, or approximately 100k steps with batch size of 64.\footnote{We use a triangular learning rate schedule with 10\% warmup and a maximum learning rate of 0.0001.}

The hyper-parameters are in Table ~\ref{tab:ptparams}. We provide code for training and evaluating our model as well.\footnote{\url{https://github.com/Kel-Lu/SciGen} \label{note:github}} Our code is based on HuggingFace's implementation of GPT2-small (117M parameters). We trained on EC2 P3.8x machines which had 4 NVidia Tesla v100 GPUs each. Both models took 24 hours to finish training.

The only hyperparameter we tune is the learning rate. We compared 1e-4 and 6e-5 for our learning rates and used validation perplexity for model selection. 

\begin{table}[h]
    \centering
    \resizebox{0.9\columnwidth}{!}{
    \begin{tabular}{lcc}
        {\bf Hyperparameter} & {\bf Pretrain } &{\bf Finetune}  \\ \hline
        Epochs &  1 & 10\\
        Effective batch size & 64 & 64\\
        Learning rate & 1e-4 & 1e-4\\
        Weight decay & 0.00 & 0.05\\
        Warmup proportion & 0.05    & 0.10 \\
    \end{tabular}}
    \caption{Hyperparameters for the further pretraining and finetuning.\vspace{-10pt}}
    \label{tab:ptparams}
\end{table}

\section{Dataset Construction}
\label{appendix:dataset}

\begin{figure*}[]
    \centering
    \includegraphics[width=0.9\textwidth]{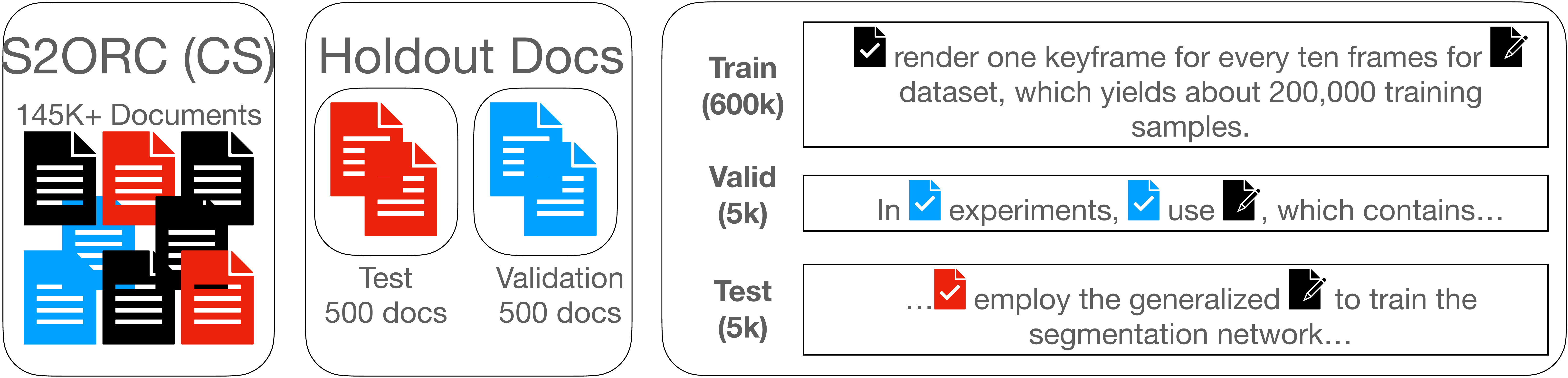}
    \caption{Dataset construction from the CS subset of S2ORC. For the far right image, we the documents with checkmarks represent the principal and those with a pencil represent the cited.\vspace{-5pt}}
    \label{fig:datasetconst}
\end{figure*}

We use data from S2ORC\footnote{\url{https://github.com/allenai/s2orc}} in both the additional pretraining and finetuning. In the former case, we use S2ORC's text with a mask over all citation references. For finetuning, we specifically train on processed data.

We process our data by extracting \source~context, cited context, and target sentence triplets. For each \source~document, we extract (1) the citation sentences, (2) the \source~document context, and (3) the citation's cited document context. We truncate the citation sentence to 100 tokens and the contexts to 450 tokens. Any remaining space is padded with a special token. The two contexts and the target citation sentence are then concatenated together with special separator tokens.

Figure~\ref{fig:datasetconst} depicts our dataset construction. To construct the data splits, we randomly select 500 \source~documents for both test and validation sets. The citation sentences that occur in these \source~documents are used as examples in the test (5310 examples) and validation (5164 examples) sets. Of the remaining examples where the principal documents were not in evaluation sets, we throw out any citation sentences that use an evaluation document as the cited document. The resultant examples are used for training (609509 examples). This construction allows us to ensure that the cited document is unseen at test time.\footnote{We also include code for data processing and splits in our repository; see footnote~\ref{note:github}.}

\section{Examples}
See Table~\ref{tab:examples}.
\begin{table*}
    \centering
    \small
    \begin{tabularx}{\textwidth}{X}
    \toprule 
         \textbf{Principal} \\
       Machine translation is important for eliminating language barriers in everyday life. To train systems which can produce good quality translations large parallel corpora are needed. Mining parallel sentences from various sources in order to train better performing MT systems is essential, especially for low resource languages. \ldots \\\vspace{-0.3em}
         \textbf{Cited} \\
        Similarity search finds application in specialized database systems handling complex data such as images or videos, which are typically represented by high-dimensional features and require specific indexing structures. This paper tackles the problem of better utilizing GPUs for this task. \ldots
        \\\vspace{-0.3em}
         \textbf{IE-based representation of cited} \\
         it \ent lopq \ent opq \ent bucket selection \ent gpu heap implementation \ldots \tfidf quantizer \tfidf memory \tfidf gemm \tfidf lane \tfidf warp \tfidf k-nn \ldots
         \\\vspace{-0.3em}
         \textbf{Sentence-based \textsc{SciGen}} \\
         For this purpose, we followed the formulation of the greedy algorithm from (\textit{Cited}) for comparing similarity lists obtained from n-best lists
         \\\vspace{-0.3em}
         \textbf{IE-based \textsc{SciGen}} \\
         In line with previous work (\textit{Cited}), we use a hash-based distance measure to calculate the similarity.
         \\\vspace{-0.3em}
         \textbf{Citing sentence} \\
          We calculate sentence similarities of each possible pairs which can be done efficiently even for large inputs (\textit{Cited}).
          \\
          \hline \hline
         \textbf{Principal} \\
       With the development of wireless technologies and popularization of smart phones, mobile traffic grown 4000-fold over the past 10 years and is expected to continue its growth at a compound annual growth rate of 53 percent from 2015 to 2020 (\textit{Cited}). The resulting problem of energy consumption on the Information and Communications Technology (ICT) has become a serious issue. \ldots \\\vspace{-0.3em}
         \textbf{Cited} \\
        We consider the problem of minimization of sum transmission energy in cellular networks where coupling occurs between cells due to mutual interference. The coupling relation is characterized by the signal-to-interference-and-noise-ratio (SINR) coupling model. Both cell load and transmission power, where cell load measures the average level of resource usage in the cell, interact via the coupling model. \ldots
        \\\vspace{-0.3em}
         \textbf{IE-based representation of cited} \\
         non-linear power coupling equation -lrb- npce -rrb- \ent non-linear load coupling equation -lrb- nlce -rrb- nlce \ent average level of usage \ldots \tfidf base \tfidf r \tfidf iap \tfidf load \tfidf cellular \ldots
         \\\vspace{-0.3em}
         \textbf{Sentence-based \textsc{SciGen}} \\
         Based on the hybrid beamforming design, the authors of (\textit{Cited}) studied the joint beamforming and power control of massive MIMO cellular networks, and proposed a multi-stage energy-efficient and fair power allocation algorithm in an RAN architecture.
         \\\vspace{-0.3em}
         \textbf{IE-based \textsc{SciGen}} \\
         In (\textit{Cited}), the load-coupled problem was addressed and the authors derived the optimal power allocation policy for the worst-case load constrained system considering the two forms of load arrival and power consumption.
         \\\vspace{-0.3em}
         \textbf{Citing sentence} \\
          Proof: The proof is similar to that of Theorem 1 in (\textit{Cited}).
          \\
          \hline \hline
         \textbf{Principal} \\
       Our lives are increasingly reliant on multimodal conversations with others. We email for business and personal purposes, attend meetings in person, chat online, and participate in blog or forum discussions. While this growing amount of personal and public conversations represent a valuable source of information, going through such overwhelming amount of data, to satisfy a particular information need, often leads to an information overload problem(\textit{Cited}). \ldots \\\vspace{-0.3em}
         \textbf{Cited} \\
         The increasing complexity of summarization systems makes it difficult to analyze exactly which modules make a difference in performance. We carried out a principled comparison between the two most commonly used schemes for assigning importance to words in the context of query focused multi-document summarization: raw frequency (word probability) and log-likelihood ratio. \ldots
        \\\vspace{-0.3em}
         \textbf{IE-based representation of cited} \\
         raw frequency \ent log-likelihood weighting scheme \ent log-likelihood ratio weighting \ent focused summarizer \ldots \tfidf 0.12717.these \tfidf topicfocused \tfidf summaries \tfidf generic \ldots
         \\\vspace{-0.3em}
         \textbf{Sentence-based \textsc{SciGen}} \\
         For generic single-query summaries, these measures often give better performance than the traditional measures (\textit{Cited}). \ldots
         \\\vspace{-0.3em}
         \textbf{IE-based \textsc{SciGen}} \\
         This score is the same as that used by (\textit{Cited}) to generate query-focused summaries from document classification.
         \\\vspace{-0.3em}
         \textbf{Citing sentence} \\
          In this work, we use log-likelihood ratio to extract the signature terms from chat logs, since log-likelihood ratio leads to better results(\textit{Cited}).
          \\
          \bottomrule
    \end{tabularx}
    \caption{Examples of system inputs and outputs from test set. }
    \label{tab:examples}
\end{table*}



\section{Human Evaluation Details}
\label{appendix:humaneval}
To ensure no annotator sees the output of more than one system on each datapoint, we randomly select 50 datapoints for each system ({\it \source~intro $\times$ cited abs}, IR, and Gold \citingsents{}) from the subset of our test data whose \source~documents appear in the ACL anthology. 
We collect judgments from 37 NLP researchers with varying levels of expertise, the majority of whom are graduate students. 
Each judge is given 15 datapoints for each of the specificity and correctness qualities.
Judges are shown a table of datapoints asked to mark whether each meets (``Yes") or fails to meet (``No") the condition. 
Judges are permitted to label ``?" or skip examples they feel uncertain about or unqualified to judge, which we ignore.

\section{Validity of Human Judgments}
\label{appendix:validity}

To test the validity of the human judgments in Section~\ref{subsec:human1}, we conduct an additional human evaluation of gold \citingsents{} paired with different kinds of mismatched inputs: (1) the correct \source~document and a random cited document, (2) the correct cited document but a random \source~document (3) random \source~and cited documents selected from ACL anthology. 
Conditions 1 and 2 allow us to see whether human judges accept sentences which align with only one or the other of the input documents; condition 3 provides a lower bound. 
We collect 107 human evaluations of correctness across these conditions, again allowing annotators to skip datapoints they are unsure of.
The results, shown in Table~\ref{tab:wrong}, indicate that human judges will sometimes accept a \citingsent as long as one of the \source~or cited documents is correct, but at a lower rate than seen in Table~\ref{tab:human1} when both documents are correct. We note that both papers in the mismatched cases are drawn from the ACL anthology, meaning there is some amount of topical coherence in their pairing. 
There is no indication from this experiment that either the \source~or cited document is a stronger influence on a judge's correctness decision, although a larger sample size might make a clear determination. 

\begin{table}[]
    \centering
    \resizebox{0.55\linewidth}{!}{%
    \begin{tabular}{lc}
    \toprule
        & {\bf Correct }  \\ \midrule
        random cited & 45.8\\
        random \source~&  46.9 \\
        both random & 17.6 \\
\bottomrule
    \end{tabular}
    }
    \caption{Correctness judgements of incorrect citing sentences (percentages). \vspace{-10pt}}
    \label{tab:wrong}
\end{table}

\section{Further Detail on Automated Metrics Results}
\label{appendix:autometrics}

We provide a more detailed report of performance on the automated metrics in Table~\ref{tab:auto} which includes all of our models. The no \source~$\times$ cited abs model uses no information from the \source~to make its retrieval decision, demonstrating the importance of relational information. 

\section{Error Analysis}
\label{appendix:error}


We investigate the reasons why gold \citingsents{} are marked incorrect in our first human evaluation in Section~\ref{sec:continuous}. One hypothesis for this gap could be that the grammar of the sentences influenced human judgment. To test this claim, one author annotated each gold example for grammatical correctness and verified these annotations with a commercial writing assistant system.\footnote{\url{https://www.grammarly.com}} We find that gold \citingsents{} with errors are more likely to be classified as incorrect (41.1\%) than those without errors (25.4\%). These results may partially explain why evaluators rated a portion of the gold sentences as incorrect.

\begin{table*}
\centering
\resizebox{0.9\textwidth}{!}{%
\begin{tabular}{lllrrrr}
\hline
&
\fieldheading{Method} & \fieldheading{Context}& \fieldheading{BLEU} & \fieldheading{Rouge-1} &  \fieldheading{Rouge-2} &\fieldheading{Rouge-L}\\
\hline
\multirow{9}{*}{Sentence-Based} & \multirow{6}{*}{{\sc SciGen}} & \source~abs $\times$ cited abs &  9.82 & 10.7& 0.6 & 8.4\\
& & \source~abs $\times$ cited intro &9.39  &    10.7 &  0.6 &  8.4 \\
& & \source~abs $\times$ cited sample &9.60  &   10.7 &  0.7  &  8.5 \\ \cline{3-7}
& & \source~intro $\times$ cited abs & 9.92 & 11.1 &  1.0 &  8.7 \\
& & \source~intro $\times$ cited intro &  9.80 &  11.1  &   1.1 &   8.8  \\
& & \source~intro $\times$ cited sampled &  9.81 &   10.9  &  0.9 &  8.7 \\
\cline{2-7}
& \multirow{3}{*}{retrieval} & \source~abs $\times$ cited abs& 9.93  &   14.2 &  0.7  &  9.7 \\
& & \hspace{1em} + MERT (BLEU)& 10.23   &  14.3 &  0.7  &  9.8 \\
& & no \source~$\times$ cited abs & 9.79 &   14.1  &  0.6  &  9.6 \\
\hline
\multirow{4}{*}{IE-based} & \multirow{3}{*}{{\sc SciGen}} 
 &\source~intro $\times$ cited tfidf & 13.17 & 15.0 & 1.3 & 12.0\\
 & & \source~abs $\times$ cited entities & 13.10 & 14.3 & 0.8 & 11.4\\
 & & \source~intro $\times$ cited entities & 13.41 & 14.7 & 1.4 & 11.8\\

\cline{2-7}
& \multirow{3}{*}{\shortstack{+Ranking}} 
& \source~intro $\times$ cited tfidf & 13.50 & 15.5 & 1.6 & 12.3\\
& &\source~abs $\times$ cited entities & 13.28 & 14.7 & 1.0 & 11.6\\
& & \source~intro $\times$ cited entities & 13.16 & 15.0 & 1.3 & 11.8\\
\hline
\end{tabular}
}
\caption{Automatic evaluation of generated texts for all of our systems.}
\label{tab:auto}
\end{table*}

\section{Analysis of Token-Wise Overlap}
\label{appendix:token}

To get a straightforward idea of how much information useful for the relationship sentence is provided by each type of context representation, we calculate the averaged percentage of token-wise overlap of the input and the gold relationship sentence, as shown in Table \ref{table:overlap}.
While a larger overlap does not guarantee a better performance, we found the best performing {\sc SciGen} systems, with or without ranking, among those using context showing largest overlaps with gold sentences.
\newcommand{\topk}{\alpha}
\begin{figure}[]
    \centering
    \includegraphics[width=0.33\textwidth]{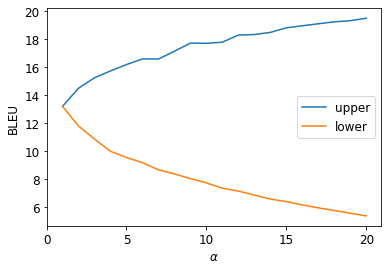}
    \caption{Upper and lower bounds of BLEU for different choices of $\topk$. \vspace{-11pt}}
    \label{fig:bounds}
\end{figure}

\begin{table*}
\centering
\resizebox{0.65\textwidth}{!}{%
\begin{tabular}{lccccc}
\hline
& \fieldheading{None} & \fieldheading{Cited abs} & \fieldheading{Cited intro} & \fieldheading{Cited tfidf} & \fieldheading{Cited entities}\\
\hline
None & N/A & 18.74 & 22.95 & 22.03 & 22.16\\
Principal abs x & 22.28 & 32.27 & 35.69 & 35.35 & 35.43\\
Principal intro x & 32.61 & 41.15 & 42.81 & 43.24 & 43.32 \\
\hline
\end{tabular}
}
\caption{\label{table:overlap} Token-wise overlap with gold relationship sentence.}
\end{table*}

\section{Oracle Study}
\label{appendix:oracle}

We conduct an oracle study to see the potential of ranking. 
Figure \ref{fig:bounds} shows the upper bound and lower bound of the BLEU score if we independently generate $\topk$ samples for each pair of $(S,C)$ using {\sc SciGen} and optimally choose the one with the highest BLEU and the one with the lowest.
With $\topk=20$, an ideal ranking system could result in a BLEU score as high as 19.50.
That provides evidence that generate-and-rank systems have the potential to surpass generate-only systems regarding BLEU.
Our proposed ranking mechanism manages to achieve higher BLEU in some cases, though it performs far below an ideal ranker.
That suggests potential future work on further improvements of ranking.

\section{Auto-Completion}
\label{appendix:autocompletion}

We notice the diversity of expressing the relationship even between the same pair of principal and cited documents.
We test whether our {\sc SciGen} could capture such diversity if provided with different triggers.
Our experiment shows that, if we provide the first three words of the relationship sentence to {\sc SciGen} and ask it to generate the rest of the sentence, the BLEU score of the generated part could be boosted to 21.38.
That suggests a use case of {\sc SciGen}, where a more personalized relationship sentence could be generated given the beginning of the sentence.

\section{Explaining without Citations}
\label{appendix:nocite}

One direction of future work is the ability to provide natural language relationship explanations to pairs of papers without a direct citation link. Table~\ref{tab:nocite} gives examples of {\sc SciGEN} output for concurrent papers at NAACL 2018. None of these papers cited each other, and thus there was no supervision in generating the explanations.

\begin{table}[ht]
    \centering
    \small
    \begin{tabularx}{\columnwidth}{lX}
    \toprule 
Principal: & \emph{Learning Joint Semantic Parsers from Disjoint Data
} \\ 
Cited: & \emph{A Transition-based Algorithm for Unrestricted AMR Parsing}\\
{\sc SciGen}: & For \emph{Principal's} task, \emph{Principal} will annotate each graph G with semantic roles from a set of annotations M, that are "required" in accordance with the AMR graph grammar \emph{Cited}.\\
\midrule
Principal: & \emph{Looking Beyond the Surface: A Challenge Set for Reading Comprehension over Multiple Sentences}  \\
Cited: & \emph{CliCR: A Dataset of Clinical Case Reports for Machine Reading Comprehension}\\
{\sc SciGen}: & For all of these datasets, \emph{Principal} focuses on performance using a common reading comprehension metric, like F1-score \emph{Cited}.\\
\midrule
Principal: & \emph{Attentive Interaction Model: Modeling Changes in View in Argumentation}  \\
Cited: & \emph{Exploring the Role of Prior Beliefs for Argument Persuasion}\\
{\sc SciGen}: & \emph{Cited} The primary strategy of this dataset is to focus on the important messages, which are important to people with different viewpoints.
\\
\bottomrule
\end{tabularx}
\caption{Example relationship explanations for pairs of papers that appeared in the same track at NAACL 2018. These papers did not cite each other. These examples have some post-processing done that replaces first person pronouns with ``\emph{principal}''.}
\label{tab:nocite}
\end{table}
\end{document}